**RESEARCH ARTICLE**

# Latent Stochastic Differential Equations for Change Point Detection

**ARTEM RYZHIKOV, MIKHAIL HUSHCHYN, AND DENIS DERKACH**
Computer Science Department, National Research University Higher School of Economics (HSE University), 10100 Moscow, Russia

Corresponding author: Artem Ryzhikov (aryzhikov@hse.ru)

This work was supported by Strategic Project "Digital Transformation: Technologies, Effects, and Performance", which is part of Higher School of Economics' development program under the "Priority 2030" academic leadership initiative. The "Priority 2030" initiative is run by Russia's Ministry of Science and Higher Education as part of National Project "Science and Universities".

**ABSTRACT** Automated analysis of complex systems based on multiple readouts remains a challenge. Change point detection algorithms are aimed to locating abrupt changes in the time series behaviour of a process. In this paper, we present a novel change point detection algorithm based on Latent Neural Stochastic Differential Equations (SDE). Our method learns a non-linear deep learning transformation of the process into a latent space and estimates a SDE that describes its evolution over time. The algorithm uses the likelihood ratio of the learned stochastic processes in different timestamps to find change points of the process. We demonstrate the detection capabilities and performance of our algorithm on synthetic and real-world datasets. The proposed method outperforms the state-of-the-art algorithms on the majority of our experiments.

## I. INTRODUCTION

Recognition of state changes of complex systems is a common task in data analysis. The system's behaviour is represented by a signal produced by continuous monitoring with multiple sensors. The task of unsupervised detection of abrupt changes in the signal forms a standalone field of research in time series analysis, and is called change point detection (CPD). CPD arises in many applications such as production quality control [1], chemical process control [2], detection of climate changes [3], human motion and health state analysis [4], aircraft monitoring [5], vibration monitoring of mechanical systems [6], seismic signal processing [7], detection of cyberattacks [8], video scene analysis [9], audio signal segmentation [10], and many others [11].

There are numerous CPD algorithms like subspace methods [12], [13], [14], [15], [16], probabilistic methods [17], [18], window-based [19], [20], clustering [21], stochastic differential equation-based methods [22], [23], etc. However, most of these methods are limited with time series dimension (and applicable to univariate time series only), change point

The associate editor coordinating the review of this manuscript and approving it for publication was Dost Muhammad Khan.

types, complexity (most of CPD algorithms are still learning-free), or robustness. Other methods, like [24] and [25], utilize deep learning (DL) power, but provide unstable results with suboptimal quality. At the same time, with the development of deep learning, most of the conventional CPD approaches remain unchanged and do not use the full power of DL.

In this work, we propose the first SDE-based likelihood-ratio CPD algorithm that utilizes the full power of DL. Namely, we use deep neural networks to learn a Latent Stochastic Differential Equation (SDE) [26], [27], which approximates the time series. The proposed method provides fitting time series dynamics, where the continuous flow is described by a Latent SDE [26], [27]. In previous works [22], [23], SDE is applied to a limited set of change points and uses a strictly limited parametric set of drift and diffusion functions. In contrast, we develop an unsupervised CPD algorithm for arbitrary-dimensional signals that uses deep learning to find the appropriate SDE and approximate its solution. The method has no limitations on the set of drift and diffusion functions or change point types. We evaluated the performance of the algorithm on open CPD benchmarks: TCPD [28], SKAB [29], TEP [30], and TSSB [21], and compared the results with available state-of-the-art methods.

The work has the following structure. Section II gives the problem statement, describes related works and background, and presents Latent SDE fit using deep neural networks. The proposed change point detection algorithm and data processing are described in Section III. Section IV defines the quality metrics that we use to compare our method with others. The experimental results and their discussion are provided in Sections V and VI respectively. Finally, the conclusion with the main results of this work is presented in Section VII.

## II. BACKGROUND, MOTIVATION, AND CHALLENGES

This section contains the CPD problem statement along with the existing CPD methods overview. We also provide here all necessary background information for our work, including the Latent SDE inference framework.

### A. PROBLEM STATEMENT

Consider a $d$-dimensional time series, where each observation at a moment $t$ is described by a vector of features $x_t \in \mathbb{R}^d$:

$$X = \{x_1, x_2, x_3, \ldots, x_\nu, x_{\nu+1}, x_{\nu+2}, \ldots\} \quad (1)$$

We assume that the distribution of all observations with $t < \nu$ are sampled from distribution $P$, and the distribution of all observations with $\nu \leq t$ are sampled from distribution $Q \neq P$. The moment $\nu \geq 1$ when the distribution changes is called a change point. In other words, the change point is the moment when a time series changes its behaviour. The illustration of several change-points is demonstrated in Fig. 1. The goal is to detect all change points in time series data. Usually, this is an unsupervised problem in statistics and machine learning due to the absence of the true positions of such points. In this work, we study this problem in such an unsupervised setting.

### B. RELATED WORK

Change point detection is a long-studied problem. The first works on the change-point detection are dated in the 50s [31], [32] for detecting a shift in the mean value of a signal for quality control of manufacturing processes. In the following decades, a range of change-point detection methods was developed that could be split into several groups based on cost function, search method, and additional constraints [33].

One of such groups is a set of subspace methods. This line of CPD algorithms is based on the time series subspace analysis of the original time series, which has a strong connection with a system identification method. This method has been thoroughly studied in control theory [12]. Some subspace methods, such as subspace identification (SI) [13] and singular spectrum transformation (SST) [15], [16], are based on classical approaches, for example, matrix and SVD transformation of the time series. Some others use more complicated neural network projections to a time-invariant subspace [24].

Another common group of change-point detection methods is based on a comparison of the empirical probability density distributions before and after change points. The CUSUM [32], [34] algorithm assumes that these distributions are known and detects a change point using a sequential hypothesis test procedure. The GLR [11], [35] method supposes that the parameters of the distribution after the change point are unknown and estimates it by likelihood minimization. Change forest [36] is a likelihood ratio classifier based on random forests that uses class probability predictions to compare different change point configurations. Generally, these methods use hypothesis comparison tests, comparing null and alternative hypothesis likelihoods in each point [37], [38], [39], [40], [41]. Usually [17], [35], [42], a null hypothesis suggests that no change points occur in the timestamp, whereas an alternative hypothesis suggests that a change point is present in the timestamp.

The next group of methods is based on estimation of some statistics. For instance, a Gaussian process (GP) is a probabilistic method to describe stationary time series analysis and prediction [18]. Another method, called Bayesian Change Point Detection (BCPD) [17], estimates the posterior distribution over an auxiliary variable run length $r_t$ which represents the time that elapsed since the last change point. Given the run length at a time instant $t$, the run length at the next time point can be either reset back to 0, if a change point occurs at this time, or increased by 1, if the current state continues for one more time unit.

A range of CPD approaches is based on direct probability densities ratio estimation for two samples without the need to know the individual densities [43]. One of the first such algorithms uses logistic regression on RBF kernels [44]. Later, other methods based on RBF kernels were proposed: KLIEP [45], uLSIF [46], and RuLSIF [47]. Their application in change-point detection are described in [12], [14], and [48].

Another group of methods also splits the time series into windows and then uses a kernel-based statistical test to assess the homogeneity between subsequent windows [19]. One of such recent deep learning kernel-based methods is called KL-CPD [25]. It uses maximum mean discrepancy (MMD) statistical test [20] between distributions $P$ and $Q$ of windows before and after the change point respectively (see previous section II-A).

One more group of CPD methods is based on cost functions [49]. These methods estimate the discrepancy score between two segments of a time series by comparing cost function values before and after splitting the segment into two by a change-point. The most popular algorithms of this group are Binseg [50], Pelt [51], [52], and Window [49].

Graph-based CPD methods first infer a graph by mapping observations (i.e., windows or sets of time series) to nodes and connecting nodes by edges if their pairwise similarity exceeds a predefined threshold. Next, a bespoke graph statistic is applied to split the graph into subgraphs leading to change points in the time series [53].

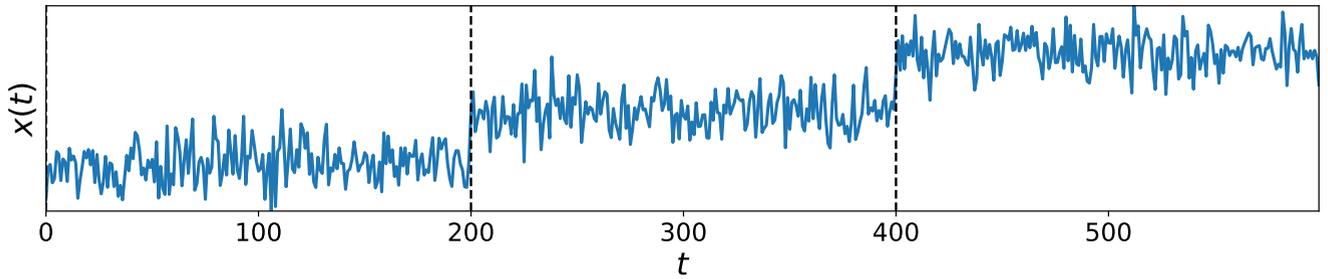

**FIGURE 1.** Example of a time series with two change points at times $v_1 = 200$ and $v_2 = 400$. At these moments, the mean value of the signal changes with jumps.

From a different perspective, the problem of change point detection can be considered as a clustering problem with a known or unknown number of clusters, such that observations within clusters are identically distributed, and observations between adjacent clusters are not. If a data point at the time stamp $t$ belongs to a different cluster than the data point at the time stamp $t + 1$, then a change point occurs between the two observations. One of such recently introduced methods is the Classification Score Profile (ClaSP) which performs segmentation of the time series [21] using KNN classification procedure.

However, most of the aforementioned CPD algorithms do not use deep learning power behind the implementation, and tend to overfit on time series outliers, or can be used in a supervised setting only. We believe that the main reason for this is the lack of labeled data in most real-world tasks, the specifics of each individual CPD case data with the resulting complexity of transfer learning, and the complexity of generalizing the aforementioned conventional approaches to a DL generalization.

In this work, we evolve conventional SDE approaches [22], [23] and propose a novel robust Latent Stochastic Differential Equation's likelihood ratio method for the change point detection problem. Unlike the previous algorithms, the proposed method combines a limited conventional time series analysis approach based on SDE with the full power of deep learning. The method is entirely unsupervised and requires no labeled change points in the training dataset.

### C. LATENT STOCHASTIC DIFFERENTIAL EQUATIONS

Consider a $D$-dimensional time series $X = \{x_t \in \mathbb{R}^D\}_{t \in \mathbb{T}}$ in time interval $\mathbb{T} = [0, T]$, $\{w_t\}_{t \in \mathbb{T}}$ is a $D$-dimensional Wiener process.

Then, a stochastic process $\{x_t\}_{t \in \mathbb{T}} \in \mathbb{R}^D$ can be defined by an Itô SDE:

$$dx_t = \mu(x_t, t)dt + \sigma(x_t, t)dw_t, \quad (2)$$

if $x_0$ is independent of the $\sigma$-algebra generated by $w_t$, and

$$x_T = x_0 + \int_0^T \mu(x_t, t)dt + \int_0^T \sigma(x_t, t)dw_t, \quad (3)$$

where, $\mu(x_t, t) : \mathbb{R}^D \times \mathbb{T} \to \mathbb{R}^D$ is the *drift function*, $\sigma(x_t, t) : \mathbb{R}^D \times \mathbb{T} \to \mathbb{R}^D$ is the *diffusion function* and the second integral in (3) is the Itô stochastic integral [54]. When the functions are globally Lipschitz, that is,

$$||\mu(x, t) - \mu(y, t)|| + ||\sigma(x, t) - \sigma(y, t)|| \leq L||x - y||$$
$$\forall x, y \in \mathbb{R}^D, t \in \mathbb{T} \quad (4)$$

for some constant $L > 0$, there exists a unique $t$-continuous strong solution to (2) [26], [54].

In case of high dimension $D$ of time series $X$, a more efficient use of SDE on latent spaces $\{z_t \in \mathbb{R}^d | d \leq D\}$ can be used [26], [27]. In [27], the authors propose an efficient variational inference framework for such latent SDE models. In particular, given observations $X$, they parameterize both a prior over functions and an approximate posterior of latent $Z$ using SDEs:

$$d\tilde{z}_t = \mu_\theta(\tilde{z}_t, t)dt + \sigma(\tilde{z}_t, t)dw_t, \quad \text{(prior)}$$
$$dz_t = \mu_\phi(z_t, t)dt + \sigma(z_t, t)dw_t, \quad \text{(approx. posterior)}$$
$$\tilde{z}_0 = z_0 \sim \psi(x_0), \quad \text{(initial latent state)}$$
$$x_t = f(z_t), \quad \text{(decoded states)}$$

where $\mu_\theta$, $\mu_\phi$, and $\sigma$ are Lipschitz in both arguments. $\mu_\theta$ is a prior drift function with prior parameters $\theta$, $\mu_\phi$ is an approximate posterior drift function with variational parameters $\phi$, $\psi$ is an encoder, and $f$ is a decoder. In such a setting, the evidence lower bound (ELBO) can be written [27] as:

$$\log p(x_1, x_2, \ldots, x_N | \theta)$$
$$\geq \mathbb{E}_{z_t} \left[ \Sigma_{i=1}^N \log p(x_{t_i} | z_{t_i}) - \int_0^T \frac{1}{2} |u(z_t, t)|^2 dt \right], \quad (5)$$

where $u(z_t, t) : \mathbb{R}^d \times \mathbb{R} \to \mathbb{R}^d$ is:

$$u(z_t, t) = \frac{\mu_\theta(z_t, t) - \mu_\phi(z_t, t)}{\sigma(z_t, t)}, \quad (6)$$

where the expectation is taken over the approximate posterior process defined by (approx. posterior). $u(z_t, t)$ can be considered as a Kullback–Leibler (KL) divergence between the approximate posterior and prior, which regularizes when the approximate posterior is far from the prior at a point $(z_t, t)$. The likelihoods of observations $x_1, \ldots, x_N$ at times $t_1, \ldots, t_N$ depend solely on latent states $z_t$ at corresponding times and a predefined likelihood function

$p(x_t|z_t)$. In practice, this function is set [26] as a Gaussian distribution:

$$p(x_t|z_t) = \mathcal{N}(x_t|f(z_t), C), \quad (7)$$

where $f$ is the decoder, $\mathcal{N}(x_t|f(z_t), C)$ is a Gaussian distribution p.d.f. with mean $f(z_t)$ and diagonal variance matrix $C$.

### D. CHANGE POINT DETECTION USING SDE
In [22], the authors propose the following SDE model of volatility change point:

$$dx_t = \mu(x_t)dt + \theta_t \sigma(x_t) dw_t \quad (8)$$

and find a time $\nu$ when the diffusion coefficient $\theta_{\nu-1} \neq \theta_\nu$ is changed. However, one of the main limitations of the method is that volatility may have more complicated kind of change than up to a multiplier $\theta$.

In [23], a more complicated volatility change point model is used:

$$dx_t = \mu(x_t)dt + \sigma(x_t; \theta)dw_t \quad (9)$$

Here, the time when the diffusion parameter $\theta$ is changed is also considered as a change point.

However, the aforementioned CPD algorithms [22], [23] still have a set of limitations. In first, they are designed to detect volatility change points only. In second, they are strictly limited with a set of drift and diffusion functions $\mu(\cdot)$, and $\sigma(\cdot; \theta)$.

For instance, in [55] the drift and diffusion functions have the following forms:

$$\mu(x) = (2 - x) \quad (10)$$
$$\sigma(x; \theta) = \theta \quad (11)$$

In [56], the forms are following:

$$\mu(x) = x \quad (12)$$
$$\sigma(x; \theta) = (1 + x^2)^\theta \quad (13)$$

## III. PROPOSED METHOD
In contrast to the previous works [22], [23], we propose a new CPD method based on Latent SDE, which approximates drift and diffusion functions with arbitrarily neural networks. The proposed method is aimed to find different types of change points in an arbitrarily multivariate time series with unknown drift and diffusion functions.

In the following section, we describe the proposed algorithm with all the related preprocessing and post-processing stages.

### A. PREPROCESSING
First, the original time series is preprocessed with a standard scaling technique. Very often, a time series has a trend and seasonality, and its observations $x_i$ can be autocorrelated. In some cases, these properties can complicate the detection of change points. To remove the trend and autocorrelation, we use the SARIMAX [57] model implemented in the Statmodels [58] package. We fit the model for the whole time series with the (5, 1, 0) order of AR parameters, differences, and MA parameters respectively. The prediction of the model for a time moment $i$ is denoted as $x_i^{SARIMAX}$. Residuals of the predictions $r_i$ are used for further analysis and are estimated as

$$r_i = x_i - x_i^{SARIMAX}. \quad (14)$$

In this work, we optionally use this preprocessing as a hyperparameter of our model for each dataset (14). Furthermore, the positional encoding features are passed as input instead of time $t$ [59].

### B. POSTPROCESSING
The output of our algorithm is multidimensional, containing the scores for all the original and auxiliary dimensions from the previous section III-A. In the post-processing stage, we use max aggregation over all the dimensions. That choice of aggregation is justified by the fact that the change point in one dimension denotes the change point of all the time series, and the most likely dimension change point at each time is taken.

Moreover, we use prominence post-processing [24] to remove the duplicated peaks around the top one.

### C. ALGORITHM
In this work, we present a novel likelihood-ratio CPD method based on latent stochastic differential equations. We use the following criterion of change point for our algorithm:

$$CPD(x_t) = \sum_{l=1}^{L} \log\left(\frac{p(x_t|t)}{p(x_t|t-l)}\right), \quad (15)$$

where $p(x|t)$ is a probability to observe $x$ at time point $t$, and $L > 0$ is a time lags range, which is considered as a hyperparameter of the algorithm. To estimate the conditional likelihood $p(x_t|\nu)$, we use the following Monte-Carlo approximation:

$$p(x_t|\nu) = \int p(x_t|z_\nu, \nu)p(z_\nu|\nu)dz \quad (16)$$
$$= \mathbb{E}_{z_\nu \sim p(z_\nu|\nu)} p(x_t|z_\nu, \nu) \quad (17)$$
$$\approx \frac{1}{N} \sum_{i=1}^{N} p(x_t|z_\nu^i, \nu), \quad (18)$$

where $N$ trajectories $z_\nu^i \sim p(z_\nu|\nu)$ are sampled from the pretrained latent SDE model [27]. The SDE dynamics of latent $Z = \{z_t\}$ is approximated by maximizing the ELBO (5) described in the previous section. If latent trajectories $z_\nu^i$ are complex enough, the likelihood function becomes $p(x_t|z_\nu^i, \nu) \approx p(x_t|z_\nu^i)$, and can be defined as in (7). So, we use the following form of CPD score at each point $x_t$:

$$CCPD(x_t) = \sum_{l=1}^{L} \log\left(\frac{p(x_t|t)}{p(x_t|t-l)}\right) \quad (19)$$

$$\approx \sum_{l=1}^{L} \log\left(\frac{\sum_{i=1}^{N} \mathcal{N}(x_t|f(z_t^i), C)}{\sum_{i=1}^{N} \mathcal{N}(x_t|f(z_{t-l}^i), C)}\right), \quad (20)$$

where $f$ is a decoder, $L = 5$ (*lags range*) and $C = 0.1 \cdot I$ (diagonal *prior variance matrix*) are hyperparameters. The final algorithm is shown in Algorithm 1.

---

**Algorithm 1** Change Point Detection With Latent SDE

---

**Input**: $D$-dimensional time series of observations $X = \{x_t \in \mathbb{R}^D\}_{t=0}^{T-1}$ at times $T = \{t_i\}_{i=0}^{T-1}$ respectively, Variational encoder $\psi$ and decoder $f$ with parameters $\Theta_\psi$ and $\Theta_f$ respectively, diffusion function $\sigma_p(z, t)$ with parameters $p$, prior drift function $\mu_\theta(z, t) = -z$, approximate posterior drift function $\mu_\phi(z, t)$ with parameters $\phi$, number of epochs $K$, learning rate $\eta$, lags range $L$, and prior variance $C$, SARIMA preprocessing option, PE is a positional encoder, $N$ is a number of trajectories to sample

1: $X = StandardScaling(X)$
2: **if** SARIMA **then**
3:    $R = X - (X)^{SARIMAX}$
4:    $X = \{X, R\}$     ▷ *Add SARIMA residuals feature(s)*
5: **end if**
6: **for** $epoch = 1, \ldots, K$ **do**
7:    Set $z_t = z_0 + \int_0^t \mu_\phi(z_v, PE(v))dv + \int_0^t \sigma_p(z_v, PE(v))dW_v\big|_{z_0=\psi(x_0)}$
8:
9:    **for** $i = 1, \ldots, T-1$ **do**
10:       Set $u(z_{t_i}) = \frac{\mu_\theta(z_{t_i}, PE(t_i)) - \mu_\phi(z_{t_i}, PE(t_i))}{\sigma_p(z_{t_i}, PE(t_i))}$
11:    **end for**
12:    $Z_t = \{z_{t_i}\}_{i=0}^{T-1}$
13:    $\Theta = \{\Theta_\psi, \Theta_f, \phi, p\}$    ▷ *Neural networks parameters*
14:    Calculate ELBO loss $\mathcal{L} = \mathbb{E}_{Z_t}\left[\sum_{i=1}^{N} \log p(x_{t_i}|z_{t_i}) - \int_0^T \frac{1}{2}|u(z_t)|^2 dt\right]$
15:    Update parameters $\Theta := \Theta + \eta \frac{\partial \mathcal{L}}{\partial \Theta}$    ▷ *any other gradient optimization can be used here*
16: **end for**
17: **for** $i = 1, \ldots, N$ **do**
18:    Sample latent trajectories $z^i \sim p(z|X)$
19: **end for**
20: **for** $i = 0, \ldots, T$ **do**
21:    $CPD(x_t) = \sum_{l=1}^{L} \log\left(\frac{\sum_{i=1}^{N} \mathcal{N}(x_t|z_t^i, C)}{\sum_{i=1}^{N} \mathcal{N}(x_t|z_{t-l}^i, C)}\right)$
22: **end for**
**Output**: change point scores for observations $X$

---

### D. THEORETICAL PROPERTIES

In this section, we provide a range of theoretical properties of the proposed algorithm for change point detection.

We demonstrate that the $CPD(x_t)$ score (15) is capable to detect changes of mean, trend, and variance of the given signal. To show that, let's estimate an approximate analytical form of the score, which is defined in Theorem 1.

*Theorem 1:* Let $v \in T$ is time moment, $L > 0$ is a predefined time lags hyperparameter (15), and $z_v$ is latent state in time $v$. Consider the following normal form of the $p(f(z_v)|v)$ and $p(x_t|f(z_v), v)$ distributions:

$$p(f(z_v)|v) = \mathcal{N}(f(z_v)|b_v, \Lambda_v), \quad (21)$$
$$p(x_t|f(z_v), v) = \mathcal{N}(x_t|f(z_v), C). \quad (22)$$

Then, the change point detection score $CPD(x_t)$ (15) takes the following analytical form:

$$CPD(x_t) =$$
$$\times \frac{1}{2} \sum_{l=1}^{L} \Big[\big(\log(|C + \Lambda_{t-l}|) + (x_t - b_{t-l})^T (C + \Lambda_{t-l})^{-1}(x_t - b_{t-l})\big) - \big(\log(|C + \Lambda_t|) + (x_t - b_t)^T (C + \Lambda_t)^{-1}(x_t - b_t)\big)\Big] \quad (23)$$

*Proof:* Let's substitute Equations (22) and (21) into the definition of $p(x_t|v)$. Since the distributions $p(x_t|f(z_v), v)$ and $p(f(z_v)|v)$ are conjugate normal, $p(x_t|v)$ is also normal:

$$p(x_t|v) = \int p(x_t|f(z), v)p(f(z)|v)df(z)$$
$$= \int \mathcal{N}(x_t|f(z), C) \times \mathcal{N}(f(z)|b_v, \Lambda_v)df(z)$$
$$= \mathcal{N}(x_t|b_v, C + \Lambda_v), \quad (24)$$

$$\log p(x_t|v) = -\frac{D}{2}\log(2\pi) - \frac{1}{2}\log(|C + \Lambda_v|)$$
$$- \frac{1}{2}(x_t - b_v)^T(C + \Lambda_v)^{-1}(x_t - b_v). \quad (25)$$

Then, the change point detection score, defined in Equation (15), takes the following form:

$$CPD(x_t) =$$
$$= \sum_{l=1}^{L} \log\left(\frac{p(x_t|t)}{p(x_t|t-l)}\right)$$
$$= \sum_{l=1}^{L} \big[\log(p(x_t|t)) - \log(p(x_t|t-l))\big]$$
$$= \frac{1}{2}\sum_{l=1}^{L} \Big[\big(\log(|C + \Lambda_{t-l}|) + (x_t - b_{t-l})^T(C + \Lambda_{t-l})^{-1}(x_t - b_{t-l})\big) - \big(\log(|C + \Lambda_t|) + (x_t - b_t)^T(C + \Lambda_t)^{-1}(x_t - b_t)\big)\Big] \quad (26)$$

$\square$

This theorem shows how the change point detection score behaves with different alternations of the signal. We assume that the mean vector $b_t$ and the covariance matrix $C + \Lambda_t$ represent the observed values. Thus, all changes of the time series are reflected in the score values.

Using Theorem (1), let's estimate the behaviour of the change point detection score for three popular cases. Consider a multivariate time series with a change point at time moment $t$. In the first case, we assume, that the mean value of the signal is changed in some time. In the second case, the trend of the signal is changed. Finally, in the third case, the variance of the signal alternates. Corollaries 1, 2, and 3 define the score values for these three change points.

*Corollary 1:* Consider a multidimensional time series with the mean jump $\Delta b$ at a time $t$. It means, that, $\Lambda_{t-l} = \Lambda_t = \Lambda$, $b_t = b + \Delta b$, and $b_{t-1} = b_{t-2} = \ldots = b_{t-L} = b$. Then, the change point detection score $CPD(x_t)$ has the following form:

$$CPD(x_t) = \frac{L}{2}\big((x_t - b)^T (C + \Lambda)^{-1} (x_t - b) \\ - (x_t - b - \Delta b)^T (C + \Lambda)^{-1} (x_t - b - \Delta b)\big). \quad (27)$$

*Corollary 2:* Consequence of (1). Consider a multidimensional time series $x$ with the trend jump $\Delta^2 b$ at the time $t$ and its first difference time series $x^\Delta$ with a constant covariance matrix $\Lambda^\Delta$ and mean jump $\Delta b^\Delta = \Delta^2 b$ at a time $t$. It means, that, $\Lambda_{t-l}^\Delta = \Lambda_t^\Delta = \Lambda^\Delta$, $b_t^\Delta = b^\Delta + \Delta^2 b$, $b_{t-1}^\Delta = b_{t-2}^\Delta = \cdots = b_{t-L}^\Delta = b^\Delta$, $x_0^\Delta = 0$, and $x_t^\Delta = x_t - x_{t-1}|t > 0$. Then, the change point detection score $CPD(x_t^\Delta)$ has the following form:

$$CPD(x_t^\Delta) \\ = \frac{L}{2}\big((x_t^\Delta - b^\Delta)^T (C + \Lambda^\Delta)^{-1} (x_t^\Delta - b^\Delta) \\ - (x_t^\Delta - b^\Delta - \Delta^2 b)^T (C + \Lambda^\Delta)^{-1} (x_t^\Delta - b^\Delta - \Delta^2 b)\big) \quad (28)$$

The Corollary (2) proves that trend changes can be also detected if first difference preprocessing is performed over the original time series $x$ and used as input to the model.

*Corollary 3:* Consider a multidimensional time series with the covariance change $\Lambda_1 \to \Lambda_2$ at a time $t$. It means, that, $b_{t-l} = b_t = b$, $\Lambda_t = \Lambda_2$, $\Lambda_{t-1} = \Lambda_{t-2} = \ldots = \Lambda_{t-L} = \Lambda_1$. Then, the change point detection score $CPD(x_t)$ has the following form:

$$CPD(x_t) = \frac{L}{2}\Big[\big(\log(|C + \Lambda_1|) - \log(|C + \Lambda_2|)\big) \\ + (x_t - b)^T \big((C + \Lambda_1)^{-1} \\ - (C + \Lambda_2)^{-1}\big)(x_t - b)\Big]. \quad (29)$$

The theorem and corollaries considered in this section provide theoretical foundation for the proposed algorithm. They demonstrate the ability of the algorithm to detect all three main types of change points on multivariate time series and predict the score values for different cases.

## IV. METRICS

Recently, some change point detection benchmarks were introduced: TCPD [28], SKAB [29], TSSB [21], and TEP [30]. In this work, we use four metrics in our experiments: *Covering* [28], *F1* [28], *Relative Change Point Distance (RCPD)* [21], [60], and *NAB* [61], that were presented in the benchmarks. This section is addressed to a detailed explanation of the change point evaluation metrics used in this work.

### A. F1

Consider a time series with several change points. Following [28], we define $\hat{\mathcal{T}} = \{\hat{\tau}_i\}_m$ as a set of change point locations provided by a detection algorithm and let $\mathcal{T} = \{\tau_i\}_n$ be a combined set of all human annotations. For a set of ground truth locations $\mathcal{T}$, they denote a set of true positives $TP = \{\tau_i | \exists \hat{\tau}_j : |\hat{\tau}_j - \tau_i| < M\}$, where $M = 5$ [28]. That means the algorithm change point prediction can be away from ground truth not farther than $M$. We also ensure that only one $\hat{\tau}_j$ can be used for a single $\tau_i$. The latter requirement is needed to avoid double-counting, and $M = 5$ is the allowed margin of error. Then, precision (P) and recall (R) are defined as,

$$P = \frac{|TP|}{|\hat{\mathcal{T}}|}, \quad R = \frac{|TP|}{|\mathcal{T}|} \quad (30)$$

Then, $F_1$ is used as a quality measure of change point detection:

$$F_1 = \frac{2PR}{P + R} \quad (31)$$

### B. COVERING

The second quality metric is based on a segmentation approach, where each change point is considered as a border between two separate segments of a time series. By analogy with other segmentation tasks, the Jaccard score can be used here:

$$J(\mathcal{A}, \mathcal{A}') = \frac{|\mathcal{A} \cap \mathcal{A}'|}{|\mathcal{A} \cup \mathcal{A}'|}, \quad (32)$$

where $\mathcal{A}$ is a ground-truth segment and $\mathcal{A}'$ is a segment formed by the found change points. To expand this metric to a multiple segments case, the authors of [28] propose the *Covering* metric:

$$C(\mathcal{G}, \mathcal{G}') = \frac{1}{T} \sum_{\mathcal{A} \in \mathcal{G}} |A| \cdot \max_{\mathcal{A}' \in \mathcal{G}'} J(\mathcal{A}, \mathcal{A}'), \quad (33)$$

where $\mathcal{G}$ and $\mathcal{G}'$ correspond to ground truth and algorithm segmentation, respectively, $T$ is a time series length. In the case of multiple annotators, the metric was averaged over the annotators labels.

### C. NAB

The Numenta Anomaly Benchmark (NAB) score was introduced in [61] and uses a distance-weighted score for predicted change points within and outside the predefined region around ground truth change points. The region starts at the ground truth position of a change point. Thus, all the change point predictions before the ground truth and all others outside the region are considered as false positives (FP). Within the region, only the nearest prediction is considered as a true positive (TP). All the rest of the predictions within the region are ignored.

For each predicted change point $\hat{\tau} \in \hat{\mathcal{T}}$, NAB score is calculated in the following way:

$$\sigma^A(\hat{\tau}) = (A_{TP} - A_{FP})\left(\frac{1}{1 + e^{5|\hat{\tau}-\tau|}}\right) - 1, \quad (34)$$

where $|\hat{\tau} - \tau|$ is a relative position of the detected $\hat{\tau}$ within the region. Here, the profile coefficients $A = \{A_{TP}, A_{FP}, A_{FN}, A_{TN}\}$ are predefined. In this work, we use 3 different profiles for the evaluation:

$$A_{Standart} = \{1.0, -0.11, -1.0, 1.0\}, \quad (35)$$
$$A_{LowFP} = \{1.0, -0.22, -1.0, 1.0\}, \quad (36)$$
$$A_{LowFN} = \{1.0, -0.11, -2.0, 1.0\}. \quad (37)$$

In our work, NAB score for these three profiles is denoted as NAB.Standart, NAB.LowFP, and NAB.LowFN respectively.

We use default region sizes in NAB scores computation. In [61], the authors show that the region size has a minor impact on the final metric value and is chosen to be in the range from $\frac{5}{|\mathcal{T}|}\%$ to $\frac{20}{|\mathcal{T}|}\%$ of the time series length $T$. For the predictions $\hat{\mathcal{T}}$, the raw score is:

$$S_{\hat{\mathcal{T}}} = \left(\sum_{\hat{\tau} \in \hat{\mathcal{T}}} \sigma^A(\hat{\tau})\right) + A_{FN}|FN|, \quad (38)$$

where $|FN|$ is a number of false negatives (empty windows with no detections around the ground truth).

Then, the final NAB score for the predictions $\hat{\mathcal{T}}$ is written in the following rescaled form:

$$\text{NAB}_{\hat{\mathcal{T}}} = 100\frac{S_{\hat{\mathcal{T}}} - S_{\text{null}}}{S_{\text{perf}} - S_{\text{null}}} \quad (39)$$

where $S_{\text{perf}}$ denotes a raw score for "perfect" detector (one that outputs all true positives and no false positives) and $S_{\text{null}}$ denotes a raw score for "null" detector (one that outputs no anomaly/change point detections).

### D. RELATIVE CHANGE POINT DISTANCE (RCPD)

RCPD was introduced in [60] and later used in the ClaSP change point algorithm [21]. It computes an average distance between the predictions $\{\hat{\tau} \in \hat{\mathcal{T}}\}$ and nearest change points $\{\tau \in \mathcal{T}\}$:

$$RCPD = \frac{1}{T|\hat{\mathcal{T}}|}\sum_{\hat{\tau} \in \hat{\mathcal{T}}} \min_{\tau \in \mathcal{T}} |\tau - \hat{\tau}|, \quad (40)$$

where $T$ is a time series length regarding the aforementioned notation in the section.

**TABLE 1.** Toy synthetic benchmark (Univariate). Each row (except header) represents the results of a specific algorithm. The first column contains algorithm names. All the rest columns correspond to specific metrics listed in the header. On each cell, a metric value averaged over all the datasets is shown. The best values for each metric are highlighted with bold. The uncertainty for each algorithm is estimated on 5 different runs.

| Algorithm | NAB (standard) | NAB (LowFP) | NAB (LowFN) | F1 | Covering | RCPD |
|---|---|---|---|---|---|---|
| Perfect detector | 100 | 100 | 100 | 1 | 1 | 0 |
| **SDE** | **93.68** ± 0.00 | **87.55** ± 0.00 | **95.78** ± 0.84 | **0.88** ± 0.0 | **0.94** ± 0.00 | **0.01** ± 0.00 |
| ClaSP | 45.98 ± 0.00 | 37.32 ± 0.00 | 49.17 ± 0.00 | 0.70 ± 0.00 | 0.72 ± 0.00 | 0.07 ± 0.00 |
| BOCPD | 88.69 ± 0.00 | 77.39 ± 0.00 | 92.47 ± 0.00 | 0.53 ± 0.00 | 0.83 ± 0.00 | 0.33 ± 0.00 |
| KLCPD | 64.30 ± 3.88 | 42.09 ± 2.86 | 74.35 ± 2.27 | 0.70 ± 0.00 | 0.84 ± 0.03 | 0.17 ± 0.00 |
| TIRE | 80.14 ± 12.61 | 73.98 ± 10.63 | 82.31 ± 11.98 | 0.73 ± 0.05 | 0.79 ± 0.06 | 0.10 ± 0.02 |
| CHANGE_FOREST | 48.67 ± 0.00 | 0.00 ± 0.00 | 94.49 ± 0.00 | 0.69 ± 0.00 | 0.78 ± 0.00 | 0.06 ± 0.00 |
| WIN | 91.73 ± 0.00 | 83.76 ± 0.00 | 94.49 ± 0.00 | 0.83 ± 0.00 | 0.83 ± 0.00 | 0.07 ± 0.00 |
| BINSEG | 92.25 ± 0.00 | 84.60 ± 0.00 | 94.83 ± 0.00 | 0.76 ± 0.00 | 0.81 ± 0.00 | 0.08 ± 0.00 |
| DYNP | 91.26 ± 0.00 | 83.23 ± 0.00 | 94.17 ± 0.00 | 0.81 ± 0.00 | 0.80 ± 0.00 | 0.14 ± 0.00 |
| KERNEL | 91.70 ± 0.00 | 83.72 ± 0.00 | 94.47 ± 0.00 | 0.83 ± 0.00 | 0.81 ± 0.00 | 0.12 ± 0.00 |
| PELT | 88.06 ± 0.00 | 76.42 ± 0.00 | 92.04 ± 0.00 | 0.81 ± 0.00 | 0.79 ± 0.00 | 0.12 ± 0.00 |
| Null detector | 0 | 0 | 0 | 0 | 0 | ∞ |

**TABLE 2.** Toy synthetic benchmark (Multivariate). Each row (except header) represents the results of a specific algorithm. The first column contains algorithm names. All the rest columns correspond to specific metrics listed in the header. On each cell, a metric value averaged over all the datasets is shown. The best values for each metric are highlighted with bold. The uncertainty for each algorithm is estimated on 5 different runs.

| Algorithm | NAB (standard) | NAB (LowFP) | NAB (LowFN) | F1 | Covering | RCPD |
|---|---|---|---|---|---|---|
| Perfect detector | 100 | 100 | 100 | 1 | 1 | 0 |
| **SDE** | **86.41** ± 0.00 | **79.91** ± 0.00 | **88.97** ± 0.88 | **0.93** ± 0.00 | **0.99** ± 0.00 | **0.01** ± 0.00 |
| ClaSP | - | - | - | - | - | - |
| BOCPD | - | - | - | - | - | - |
| KLCPD | 62.24 ± 5.62 | 47.33 ± 4.95 | 68.82 ± 4.57 | 0.73 ± 0.02 | 0.83 ± 0.02 | 0.23 ± 0.03 |
| TIRE | 75.93 ± 8.72 | 69.88 ± 7.45 | 77.95 ± 9.40 | 0.76 ± 0.04 | 0.88 ± 0.02 | 0.08 ± 0.04 |
| CHANGE_FOREST | 69.65 ± 0.00 | 49.29 ± 0.00 | 76.43 ± 0.00 | 0.90 ± 0.00 | 0.91 ± 0.00 | 0.05 ± 0.00 |
| WIN | 77.09 ± 0.00 | 55.43 ± 0.00 | 84.72 ± 0.00 | 0.70 ± 0.00 | 0.76 ± 0.00 | 0.25 ± 0.00 |
| BINSEG | 54.83 ± 0.00 | 45.36 ± 0.00 | 59.88 ± 0.00 | 0.77 ± 0.00 | 0.80 ± 0.00 | 0.18 ± 0.00 |
| DYNP | 80.80 ± 0.00 | 61.80 ± 0.00 | 87.20 ± 0.00 | 0.73 ± 0.00 | 0.81 ± 0.00 | 0.15 ± 0.00 |
| KERNEL | 81.51 ± 0.00 | 73.23 ± 0.00 | 84.34 ± 0.00 | 0.80 ± 0.00 | 0.81 ± 0.00 | 0.11 ± 0.00 |
| PELT | 0.00 ± 0.00 | 0.00 ± 0.00 | 0.00 ± 0.00 | 0.67 ± 0.00 | 0.76 ± 0.00 | 0.14 ± 0.00 |
| Null detector | 0 | 0 | 0 | 0 | 0 | ∞ |

## V. EXPERIMENTAL RESULTS

In this section, we describe an evaluation of our model along with a comparison of state-of-the-art CPD algorithms like KL-CPD [25], TIRE [24], ClaSP [21], BOCPD [17], BINSEG [62], CHANGE_FOREST [36] and *ruptures* [49] algorithms like PELT, OPT, KERNEL, and WIN. Each algorithm is evaluated at the best threshold for each quality metric described in the previous section. If the algorithm does not return detection scores, the optimal number of the predicted change points is taken for each metric. Univariate algorithms, like ClaSP and BOCPD are evaluated on univariate datasets only. For uncertainty estimation, each algorithm was trained and evaluated on each dataset 5 times from different initialization seeds.

In further subsections, we describe all the evaluation corpuses in detail and provide aggregated results tables over each corpus. A detailed models and hyperparameters setup is described in Appendix VII-A. All other implementation details and datasets are available at our public paper repositories for source code[1] and data[2] respectively.

### A. OUR SYNTHETIC STUDIES

To test the model performance, we first take a simple set of synthetic experiments using artificial datasets. The aim of this synthetic experiment is to check how well our algorithm detects different kinds of change points: trend changes, mean jumps, and volatility changes on univariate and multivariate cases. Each experiment represents noisy data generation with further models quality and robustness estimation.

---
[1] https://gitlab.com/lambda-hse/neural-sde-for-cpd
[2] https://gitlab.com/lambda-hse/change-point/datasets/

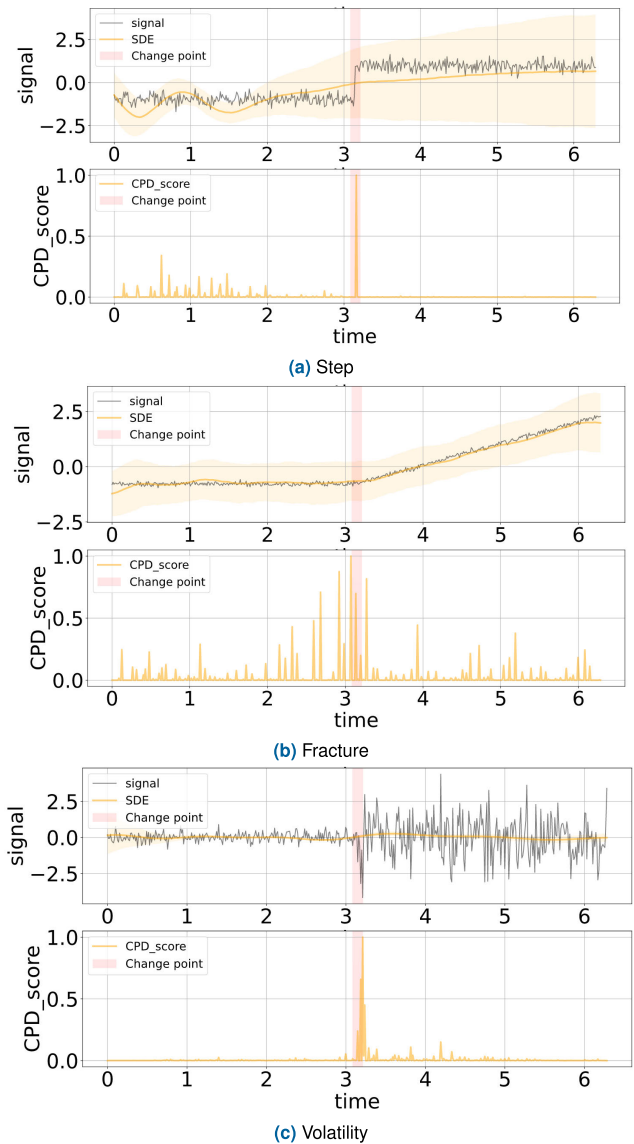

**FIGURE 2.** In these figures, the behaviour of our algorithm on three main types of change points is shown. On (a), the mean change point (step) is shown. On (b), the trend change point (fracture) is figured. On (c), the volatility change point is shown. The top halves of the figures contain original time series with averaged SDE dynamics. The bottom halves of the figures represent the score of our algorithm, maximized over all the original and auxiliary dimensions.

Figure 2 shows the behaviour of the algorithm on 3 simulated change point types: mean, trend, and volatility change points.

The averaged metric values for synthetic corpus are listed in Tables 1,2. For univariate datasets, our algorithm outperforms all the others on all CP metrics. For multivariate datasets, the algorithm outperforms the others on all metrics besides *Low.FN*.

**TABLE 3.** Results on univariate TCPD datasets [28]. Each row (except header) represents the results of a specific algorithm. The first column contains algorithm names. All the rest columns correspond to specific metrics listed in the header. On each cell, a metric value averaged over all the datasets is shown. The best values for each metric are highlighted in bold. The uncertainty for each algorithm is estimated on 5 different runs.

| Algorithm | NAB (standard) | NAB (LowFP) | NAB (LowFN) | F1 | Covering | RCPD |
|---|---|---|---|---|---|---|
| Perfect detector | 100 | 100 | 100 | 1 | 1 | 0 |
| **SDE** | **96.43** ± 0.11 | **92.95** ± 0.23 | **97.62** ± 0.07 | **0.77** ± 0.01 | **0.86** ± 0.00 | **0.03** ± 0.01 |
| ClaSP | 46.45 ± 0.00 | 37.01 ± 0.00 | 52.2 ± 0.00 | 0.34 ± 0.00 | 0.58 ± 0.00 | 0.20 ± 0.00 |
| BOCPD | 87.81 ± 0.00 | 81.64 ± 0.00 | 91.31 ± 0.00 | 0.45 ± 0.00 | 0.74 ± 0.00 | 0.08 ± 0.00 |
| KLCPD | 62.79 ± 1.52 | 50.00 ± 1.95 | 69.95 ± 0.94 | 0.52 ± 0.02 | 0.71 ± 0.01 | 0.15 ± 0.01 |
| TIRE | 59.55 ± 9.93 | 57.07 ± 9.56 | 63.13 ± 8.28 | 0.46 ± 0.04 | 0.70 ± 0.04 | 0.15 ± 0.03 |
| CHANGE_FOREST | 25.12 ± 0.00 | 0.00 ± 0.00 | 40.60 ± 0.00 | 0.48 ± 0.00 | 0.63 ± 0.00 | 0.14 ± 0.00 |
| WIN | 86.68 ± 0.00 | 80.38 ± 0.00 | 88.99 ± 0.00 | 0.55 ± 0.00 | 0.81 ± 0.00 | 0.12 ± 0.00 |
| BINSEG | 91.63 ± 0.00 | 84.91 ± 0.00 | 93.96 ± 0.00 | 0.67 ± 0.00 | 0.81 ± 0.00 | 0.06 ± 0.00 |
| OPT | 90.30 ± 0.00 | 82.29 ± 0.00 | 93.27 ± 0.00 | 0.6 ± 0.00 | 0.72 ± 0.00 | 0.11 ± 0.00 |
| KERNEL | 88.23 ± 0.00 | 79.19 ± 0.00 | 91.89 ± 0.00 | 0.54 ± 0.00 | 0.73 ± 0.00 | 0.14 ± 0.00 |
| PELT | 34.86 ± 0.00 | 0.00 ± 0.00 | 56.31 ± 0.00 | 0.54 ± 0.00 | 0.64 ± 0.00 | 0.14 ± 0.00 |
| Null detector | 0 | 0 | 0 | 0 | 0 | ∞ |

**TABLE 4.** Results on multivariate TCPD datasets [28]. Each row (except header) represents the results of a specific algorithm. The first column contains algorithm names. All the rest columns correspond to specific metrics listed in the header. On each cell, a metric value averaged over all the datasets is shown. The best values for each metric are highlighted in bold. The uncertainty for each algorithm is estimated on 5 different runs.

| Algorithm | NAB (standard) | NAB (LowFP) | NAB (LowFN) | F1 | Covering | RCPD |
|---|---|---|---|---|---|---|
| Perfect detector | 100 | 100 | 100 | 1 | 1 | 0 |
| **SDE** | **94.29** ± 0.12 | **89.19** ± 0.22 | **96.19** ± 0.07 | **0.66** ± 0.01 | **0.85** ± 0.00 | **0.01** ± 0.01 |
| ClaSP | - | - | - | - | - | - |
| BOCPD | - | - | - | - | - | - |
| KLCPD | 69.79 ± 6.31 | 52.98 ± 10.37 | 78.44 ± 4.85 | 0.40 ± 0.01 | 0.67 ± 0.04 | 0.07 ± 0.05 |
| TIRE | 63.18 ± 14.94 | 59.91 ± 12.80 | 64.45 ± 16.83 | 0.45 ± 0.06 | 0.63 ± 0.10 | 0.12 ± 0.03 |
| CHANGE_FOREST | 42.59 ± 0.00 | 0.00 ± 0.00 | 61.22 ± 0.00 | 0.46 ± 0.00 | 0.55 ± 0.00 | 0.06 ± 0.00 |
| WIN | 87.77 ± 0.00 | 84.27 ± 0.00 | 89.32 ± 0.00 | 0.55 ± 0.00 | 0.75 ± 0.00 | 0.03 ± 0.00 |
| BINSEG | 90.08 ± 0.00 | 81.79 ± 0.00 | 92.88 ± 0.00 | 0.56 ± 0.00 | 0.76 ± 0.00 | 0.06 ± 0.00 |
| OPT | 89.18 ± 0.00 | 81.5 ± 0.00 | 91.79 ± 0.00 | 0.55 ± 0.00 | 0.76 ± 0.00 | 0.06 ± 0.00 |
| KERNEL | 91.21 ± 0.00 | 84.05 ± 0.00 | 93.63 ± 0.00 | 0.57 ± 0.00 | 0.71 ± 0.00 | 0.05 ± 0.00 |
| PELT | 20.20 ± 0.00 | 0.00 ± 0.00 | 0.00 ± 0.00 | 0.51 ± 0.00 | 0.64 ± 0.00 | 0.11 ± 0.00 |
| Null detector | 0 | 0 | 0 | 0 | 0 | ∞ |

**TABLE 5.** Results on SKAB benchmark datasets [29]. Each row (except header) represents the results of a specific algorithm. The first column contains algorithm names. All the rest columns correspond to specific metrics listed in the header. On each cell, a metric value averaged over all the datasets is shown. The best values for each metric are highlighted in bold. The uncertainty for each algorithm is estimated on 5 different runs.

| Algorithm | NAB (standard) | NAB (LowFP) | NAB (LowFN) | F1 | Covering | RCPD |
|---|---|---|---|---|---|---|
| Perfect detector | 100 | 100 | 100 | 1 | 1 | 0 |
| **SDE** | **91.08** ± 0.13 | **84.91** ± 0.32 | **93.54** ± 0.08 | **0.50** ± 0.01 | **0.77** ± 0.00 | **0.05** ± 0.00 |
| ClaSP | - | - | - | - | - | - |
| BOCPD | - | - | - | - | - | - |
| KLCPD | 56.06 ± 3.05 | 40.69 ± 3.31 | 65.17 ± 2.77 | 0.39 ± 0.00 | 0.65 ± 0.02 | 0.14 ± 0.02 |
| TIRE | 56.45 ± 6.58 | 49.92 ± 5.86 | 63.14 ± 5.93 | 0.41 ± 0.01 | 0.70 ± 0.01 | 0.15 ± 0.01 |
| CHANGE_FOREST | 15.71 ± 0.00 | 0.00 ± 0.00 | 43.56 ± 0.00 | 0.38 ± 0.00 | 0.63 ± 0.00 | 0.15 ± 0.00 |
| WIN | 84.49 ± 0.00 | 72.19 ± 0.00 | 89.41 ± 0.00 | 0.46 ± 0.00 | 0.65 ± 0.00 | 0.12 ± 0.00 |
| BINSEG | 85.53 ± 0.00 | 73.81 ± 0.00 | 90.02 ± 0.00 | 0.44 ± 0.00 | 0.71 ± 0.00 | 0.10 ± 0.00 |
| OPT | 83.24 ± 0.00 | 70.08 ± 0.00 | 88.34 ± 0.00 | 0.44 ± 0.00 | 0.68 ± 0.00 | 0.11 ± 0.00 |
| KERNEL | 85.27 ± 0.00 | 72.97 ± 0.00 | 89.94 ± 0.00 | 0.42 ± 0.00 | 0.61 ± 0.00 | 0.17 ± 0.00 |
| PELT | 0.00 ± 0.00 | 0.00 ± 0.00 | 0.00 ± 0.00 | 0.40 ± 0.00 | 0.63 ± 0.00 | 0.13 ± 0.00 |
| Null detector | 0 | 0 | 0 | 0 | 0 | ∞ |

The complete description of the synthetic corpus, along with detailed case studies is listed in Appendix VII-B.

### B. EVALUATION ON OPEN DATASETS

In this section, we evaluate our algorithm and compare it with baselines on open datasets for change point detection: TCPD [28], SKAB [29], TSSB [21], and Tennessee Eastman Process(TEP) [30].

TCPD dataset consists of 37 real time series collected for the change point detection benchmark [28]. It includes 33 univariate and 4 multivariate series with manually labeled change points.

The SKAB corpus contains 35 individual data files. Each file represents a single experiment and contains a single anomaly (change point). The dataset represents a multivariate time series collected from the sensors installed on the testbed [29].

**TABLE 6.** Results on TSSB benchmark datasets [21]. Each row (except header) represents the results of a specific algorithm. The first column contains algorithm names. All the rest columns correspond to specific metrics listed in the header. On each cell, a metric value averaged over all the datasets is shown. The best values for each metric are highlighted in bold. The uncertainty for each algorithm is estimated on 5 different runs.

| Algorithm | NAB (standard) | NAB (LowFP) | NAB (LowFN) | F1 | Covering | RCPD |
|---|---|---|---|---|---|---|
| Perfect detector | 100 | 100 | 100 | 1 | 1 | 0 |
| **SDE** | 73.75 ± 1.06 | 61.91 ± 1.01 | 80.20 ± 0.97 | **0.62 ± 0.00** | 0.81 ± 0.00 | 0.08 ± 0.00 |
| ClaSP | 53.29 ± 0.00 | 46.81 ± 0.00 | 56.22 ± 0.00 | 0.60 ± 0.00 | 0.83 ± 0.00 | 0.02 ± 0.00 |
| BOCPD | 73.64 ± 0.00 | 58.98 ± 0.00 | 80.24 ± 0.00 | 0.50 ± 0.00 | 0.67 ± 0.00 | 0.22 ± 0.00 |
| KLCPD | 69.55 ± 1.12 | 55.99 ± 1.55 | 77.26 ± 0.92 | 0.51 ± 0.00 | 0.68 ± 0.01 | 0.24 ± 0.01 |
| TIRE | 84.11 ± 1.91 | 73.87 ± 2.06 | 87.86 ± 2.06 | 0.54 ± 0.00 | 0.70 ± 0.01 | 0.14 ± 0.00 |
| CHANGE_FOREST | 0.00 ± 0.00 | 0.00 ± 0.00 | 0.00 ± 0.00 | 0.54 ± 0.00 | 0.71 ± 0.00 | 0.17 ± 0.00 |
| WIN | 77.58 ± 0.00 | 63.58 ± 0.00 | 84.42 ± 0.00 | 0.51 ± 0.00 | 0.66 ± 0.00 | 0.17 ± 0.00 |
| BINSEG | 34.78 ± 0.00 | 24.79 ± 0.00 | 42.95 ± 0.00 | 0.50 ± 0.00 | 0.47 ± 0.00 | 0.32 ± 0.00 |
| OPT | 62.30 ± 0.00 | 47.83 ± 0.00 | 70.09 ± 0.00 | 0.50 ± 0.00 | 0.63 ± 0.00 | 0.23 ± 0.00 |
| KERNEL | 65.45 ± 0.00 | 50.34 ± 0.00 | 72.89 ± 0.00 | 0.54 ± 0.00 | 0.69 ± 0.00 | 0.16 ± 0.00 |
| PELT | 0.00 ± 0.00 | 0.00 ± 0.00 | 0.00 ± 0.00 | 0.50 ± 0.00 | 0.66 ± 0.00 | 0.17 ± 0.00 |
| Null detector | 0 | 0 | 0 | 0 | 0 | ∞ |

**TABLE 7.** Results on Tennessee Eastman Process datasets [30]. Each row (except header) represents the results of a specific algorithm. The first column contains algorithm names. All the rest columns correspond to specific metrics listed in the header. On each cell, a metric value averaged over all the datasets is shown. The best values for each metric are highlighted with bold. The uncertainty for each algorithm is estimated on 5 different runs and 21 data samples.

| Algorithm | NAB (standard) | NAB (LowFP) | NAB (LowFN) | F1 | Covering | RCPD |
|---|---|---|---|---|---|---|
| Perfect detector | 100 | 100 | 100 | 1 | 1 | 0 |
| **SDE** | 94.31 ± 0.55 | 88.62 ± 2.81 | 96.2 ± 0.62 | **0.8 ± 0.16** | **0.97 ± 0.03** | **0.02 ± 0.03** |
| ClaSP | - | - | - | - | - | - |
| BOCPD | - | - | - | - | - | - |
| KL-CPD | 91.2 ± 13.15 | 83.33 ± 19.59 | 94.13 ± 6.62 | 0.67 ± 0.03 | 0.77 ± 0.08 | 0.37 ± 0.24 |
| TIRE | 58.78 ± 46.52 | 54.65 ± 41.94 | 60.19 ± 47.30 | 0.67 ± 0.06 | 0.72 ± 0.10 | 0.17 ± 0.16 |
| CHANGE_FOREST | 0.00 ± 0.00 | 0.00 ± 0.00 | 0.00 ± 0.00 | 0.67 ± 0.00 | 0.77 ± 0.07 | 0.35 ± 0.19 |
| BINSEG | 91.64 ± 4.27 | 84.59 ± 6.10 | 94.42 ± 2.73 | 0.68 ± 0.09 | 0.80 ± 0.09 | 0.40 ± 0.31 |
| DYNP | 91.71 ± 3.98 | 83.7 ± 9.76 | 94.47 ± 3.79 | 0.69 ± 0.09 | 0.81 ± 0.10 | 0.28 ± 0.26 |
| KERNEL | 91.71 ± 3.26 | 83.7 ± 5.86 | 94.47 ± 2.11 | 0.69 ± 0.07 | 0.81 ± 0.08 | 0.28 ± 0.25 |
| WIN | 93.08 ± 2.87 | 86.49 ± 4.38 | 95.38 ± 1.14 | 0.67 ± 0.07 | 0.80 ± 0.10 | 0.35 ± 0.29 |
| PELT | 0 ± 0.00 | 0 ± 0.00 | 0 ± 0.00 | 0.67 ± 0.00 | 0.88 ± 0.00 | 0.11 ± 0.00 |
| Null detector | 0 | 0 | 0 | 0 | 0 | ∞ |

**TABLE 8.** List of synthetic datasets. The first column contains dataset indices. The second column contains types of change points which presents in a dataset. The last column contains detailed dataset description.

| Dataset index | Change point types | Dataset description |
|---|---|---|
| 1 | volatility | Gumbel component is changed (2nd component is Gaussian) |
| 2 | volatility | Both Gumbel components are changed |
| 3 | volatility | Gumbel component is changed (2nd component is Gumbel) |
| 4 | volatility | Covariance of multivariate 2D Gaussian is changed |
| 5 | volatility | Single fracture in both components (2D) |
| 6 | volatility | 2 components of 2D Gaussian noise are changed |
| 7 | volatility | 1 component of 2D Gaussian noise is changed |
| 8 | trend | Single fracture in 1 component (2nd component is flat) |
| 9 | trend | Single fracture (1D) |
| 10 | trend, jump | Multiple steps and fractures (1D) |
| 11 | jump | Single step in 1 component (2nd component is flat) |
| 12 | jump | Single step (1D) |
| 13 | jump | Single step in both components |

TSSB benchmark consists of 98 univariate datasets: 49 single change point time series, 22 datasets with two change points each, 10 datasets with three change points each, and 11 datasets with 4 change points each [21].

TEP benchmark contains 22 53-dimensional single change point datasets [30].

For univariate and multivariate datasets of the TCPD corpus, the results are shown in Tables 3,4 respectively. The results for the SKAB, TSSB, and TEP corpuses are shown in Tables 5,6,7 respectively.

On the TSSB benchmark, the proposed algorithm outperforms the others only on the *F1* score. However, in all the rest studied benchmarks, our algorithm outperforms all other algorithms.

## VI. DISCUSSION

Theorem 1 and the corresponding corollaries 1, 2, 3 prove the theoretical discriminating power on trend, jump,

```
NeuralSDE(
  (approximate posterior drift): DNN(
    (net): Sequential(
      (0): Linear(
        in_features=n_pos_encodings+2*D,
        out_features=200, bias=True)
      (1): Tanh()
      (2): Linear(in_features=200,
        out_features=200, bias=True)
      (3): Tank()
      (4): Linear(in_features=200,
        out_features=2*D, bias=True)
    )
  )
  (shared diffusion): 1 (constant)
)
```

**FIGURE 3.** Architecture of Neural SDE for CPD. In our work, we use 3-layer dense neural network to approximate posterior drift, and constant diffusion (equal to 1 in each point). This configuration is similar to the default configuration in the original neural SDE work [27]. Here, *D* denotes the dimensionality of original dataset, *n_pos_encodings* denotes the number of the added time positional encoding features. Factor 2 before the dimensionality *D* means that we augment each time series with SARIMAX residuals of the same dimensionality.

**TABLE 9.** NAB.Stadart metric for synthetic datasets [61]. The first column contains dataset indices. The header contains algorithm names. In cells, NAB.Stadart metric for the corresponding dataset and algorithm is given.

| Dataset | SDE | ClaSP | BOCPD | KLCPD | TIRE | CHANGE_FOREST | WIN | BINSEG | OPT | KERNEL | PELT |
|---|---|---|---|---|---|---|---|---|---|---|---|
| 1 | 94.50 | - | - | 89.00 | 70.88 | 89.00 | 78.00 | 17.49 | 45.00 | 94.50 | 0.00 |
| 2 | 0.00 | - | - | 39.17 | 89.00 | 89.00 | 45.00 | 0.00 | 72.50 | 0.00 | 0.00 |
| 3 | 94.50 | - | - | 23.50 | 94.50 | 0.00 | 51.12 | 0.00 | 83.50 | 89.00 | 0.00 |
| 4 | 93.07 | - | - | 0.00 | 94.50 | 89.00 | 55.98 | 0.00 | 61.50 | 89.00 | 0.00 |
| 5 | 97.13 | - | - | 68.82 | 70.77 | 31.23 | 90.19 | 91.64 | 88.90 | 94.41 | 80.73 |
| 6 | 94.50 | - | - | 87.17 | 89.00 | 89.00 | 94.50 | 93.80 | 94.50 | 94.50 | 0.00 |
| 7 | 94.50 | - | - | 72.76 | 89.00 | 89.00 | 91.83 | 92.21 | 92.38 | 86.71 | 0.00 |
| 8 | 97.23 | - | - | 84.74 | 75.52 | 47.72 | 86.25 | 91.64 | 91.71 | 94.46 | 0.00 |
| 9 | 94.50 | 94.50 | 89.00 | 31.25 | 94.50 | 0.00 | 94.50 | 94.50 | 93.07 | 94.44 | 83.49 |
| 10 | 90.97 | 48.94 | 88.08 | 50.28 | 66.15 | 61.50 | 91.70 | 93.25 | 91.70 | 91.67 | 91.70 |
| 11 | 89.00 | - | - | 93.40 | 35.60 | 89.00 | 89.00 | 89.00 | 89.00 | 89.00 | 0.00 |
| 12 | 94.50 | 0.00 | 89.00 | 92.30 | 89.00 | 89.00 | 89.00 | 89.00 | 89.00 | 89.00 | 89.00 |
| 13 | 89.00 | - | - | 93.12 | 44.50 | 89.00 | 89.00 | 89.00 | 89.00 | 89.00 | 89.00 |

**TABLE 10.** NAB.LowFP metric for synthetic datasets [61]. The first column contains dataset indices. The header contains algorithm names. In cells, NAB.LowFP metric for the corresponding dataset and algorithm is given.

| Dataset name | SDE | ClaSP | BOCPD | KLCPD | TIRE | CHANGE_FOREST | WIN | BINSEG | OPT | KERNEL | PELT |
|---|---|---|---|---|---|---|---|---|---|---|---|
| 1 | 89.00 | - | - | 78.00 | 66.75 | 78.00 | 56.00 | 0.00 | 0.00 | 89.00 | 0.00 |
| 2 | 0.00 | - | - | 11.67 | 89.00 | 78.00 | 0.00 | 0.00 | 45.00 | 0.00 | 0.00 |
| 3 | 89.00 | - | - | 7.00 | 89.00 | 0.00 | 6.63 | 0.00 | 67.00 | 78.00 | 0.00 |
| 4 | 87.42 | - | - | 0.00 | 89.00 | 78.00 | 11.97 | 0.00 | 23.00 | 78.00 | 0.00 |
| 5 | 94.36 | - | - | 37.86 | 66.63 | 0.00 | 84.26 | 83.44 | 77.89 | 88.90 | 61.48 |
| 6 | 89.00 | - | - | 74.33 | 78.00 | 78.00 | 89.00 | 88.23 | 89.00 | 89.00 | 0.00 |
| 7 | 89.00 | - | - | 59.73 | 78.00 | 78.00 | 86.06 | 86.48 | 86.67 | 75.48 | 0.00 |
| 8 | 94.48 | - | - | 72.33 | 71.11 | 74.42 | 83.44 | 83.45 | 88.96 | 0.00 | 0.00 |
| 9 | 89.00 | 89.00 | 78.00 | 0.00 | 89.00 | 0.00 | 89.00 | 89.00 | 87.42 | 88.93 | 66.99 |
| 10 | 82.64 | 33.97 | 76.17 | 29.98 | 62.18 | 23.00 | 84.27 | 86.80 | 84.27 | 84.24 | 84.27 |
| 11 | 78.00 | - | - | 86.80 | 31.20 | 78.00 | 78.00 | 78.00 | 78.00 | 78.00 | 0.00 |
| 12 | 89.00 | 0.00 | 78.00 | 84.60 | 78.00 | 78.00 | 78.00 | 78.00 | 78.00 | 78.00 | 78.00 |
| 13 | 78.00 | - | - | 86.25 | 39.00 | 78.00 | 78.00 | 78.00 | 78.00 | 78.00 | 78.00 |

**TABLE 11.** NAB.LowFN metric for synthetic datasets [61]. The first column contains dataset indices. The header contains algorithm names. In cells, NAB.LowFN metric for the corresponding dataset and algorithm is given.

| Dataset name | SDE | ClaSP | BOCPD | KLCPD | TIRE | CHANGE_FOREST | WIN | BINSEG | OPT | KERNEL | PELT |
|---|---|---|---|---|---|---|---|---|---|---|---|
| 1 | **96.33** | - | - | 92.67 | 72.25 | 92.67 | 85.33 | 44.99 | 63.33 | 96.33 | 0.00 |
| 2 | 7.67 | - | - | 48.33 | 96.33 | 92.67 | 63.33 | 0.00 | 81.67 | 0.00 | 0.00 |
| 3 | **96.33** | - | - | 30.78 | 96.33 | 0.00 | 67.41 | 0.00 | 89.00 | 92.67 | 0.00 |
| 4 | 95.38 | - | - | 0.00 | 96.33 | 92.67 | 70.65 | -3.67 | 74.33 | 92.67 | 0.00 |
| 5 | **98.08** | - | - | 79.21 | 72.18 | 54.15 | 93.46 | 94.43 | 92.60 | 96.27 | 87.15 |
| 6 | **96.33** | - | - | 91.45 | 92.67 | 92.67 | 96.33 | 95.86 | 96.33 | 96.33 | 0.00 |
| 7 | **96.33** | - | - | 81.84 | 92.67 | 92.67 | 94.55 | 94.81 | 94.92 | 91.14 | 0.00 |
| 8 | **98.16** | - | - | 89.83 | 77.02 | 65.15 | 90.83 | 94.42 | 94.47 | 96.31 | 32.17 |
| 9 | **96.33** | 96.33 | 92.67 | 54.17 | 96.33 | 30.33 | 96.33 | 96.33 | 95.38 | 96.29 | 89.00 |
| 10 | 93.98 | 54.85 | 92.06 | 61.30 | 67.71 | 74.33 | 94.46 | 95.50 | 94.46 | 94.45 | 94.46 |
| 11 | 92.67 | - | - | 95.60 | 37.06 | 92.67 | 92.67 | 92.67 | 92.67 | 92.67 | 1.00 |
| 12 | **96.33** | 0.00 | 92.67 | 94.87 | 92.67 | 92.67 | 92.67 | 92.67 | 92.67 | 92.67 | 92.67 |
| 13 | 92.67 | - | - | 95.42 | 46.34 | 92.67 | 92.67 | 92.67 | 92.67 | 92.67 | 92.67 |

and volatility change points under the certain conditions (21), (22). This provides a strong hint for good phenomenological results. The obtained experimental results support the aforementioned theoretical properties of the proposed algorithm (follow the Appendix for more details and experiments). According to the experiments and theoretical properties, the algorithm effectively detects all the main types of change points and outperforms the existing CPD methods on a wide range of benchmarks. Unlike the previous algorithms, our method shows the best average quality both on univariate and multivariate datasets.

**TABLE 12.** F1 metric for synthetic datasets [28]. The first column contains dataset indices. The header contains algorithm names. In cells, F1 metric for the corresponding dataset and algorithm is given.

| Dataset name | SDE | ClaSP | BOCPD | KLCPD | TIRE | CHANGE_FOREST | WIN | BINSEG | OPT | KERNEL | PELT |
|---|---|---|---|---|---|---|---|---|---|---|---|
| 1 | **1.00** | - | - | 0.78 | 0.83 | 1.00 | 0.67 | 1.00 | 1.00 | 0.67 | 0.67 |
| 2 | **1.00** | - | - | 0.67 | 0.67 | 1.00 | 0.67 | 1.00 | 0.67 | 1.00 | 0.67 |
| 3 | 0.80 | - | - | 0.67 | 0.79 | 1.00 | 0.67 | 0.67 | 0.67 | 0.67 | 0.67 |
| 4 | 0.67 | - | - | 0.67 | 1.00 | 1.00 | 0.67 | 0.67 | 0.67 | 0.67 | 0.67 |
| 5 | **1.00** | - | - | 0.50 | 0.50 | 0.50 | 0.50 | 0.50 | 0.50 | 0.50 | 0.50 |
| 6 | **1.00** | - | - | 0.67 | 0.67 | 1.00 | 0.67 | 0.67 | 0.67 | 1.00 | 0.67 |
| 7 | **1.00** | - | - | 0.67 | 0.67 | 1.00 | 0.67 | 0.67 | 0.67 | 1.00 | 0.67 |
| 8 | 0.80 | - | - | 0.50 | 0.50 | 0.50 | 0.50 | 0.50 | 0.50 | 0.50 | 0.50 |
| 9 | 0.67 | 1.00 | 0.67 | 0.67 | 0.67 | 0.67 | 0.67 | 0.67 | 0.67 | 0.67 | 0.67 |
| 10 | **0.83** | 0.44 | 0.25 | 0.44 | 0.52 | 0.41 | 0.83 | 0.62 | 0.77 | 0.83 | 0.77 |
| 11 | **1.00** | - | - | 1.00 | 0.80 | 1.00 | 1.00 | 1.00 | 1.00 | 1.00 | 0.67 |
| 12 | **1.00** | 0.67 | 0.67 | 1.00 | 1.00 | 1.00 | 1.00 | 1.00 | 1.00 | 1.00 | 1.00 |
| 13 | **1.00** | - | - | 1.00 | 0.83 | 1.00 | 1.00 | 1.00 | 1.00 | 1.00 | 1.00 |

**TABLE 13.** Covering metric for synthetic datasets [28]. The first column contains dataset indices. The header contains algorithm names. In cells, Covering metric for the corresponding dataset and algorithm is given.

| Dataset name | SDE | ClaSP | BOCPD | KLCPD | TIRE | CHANGE_FOREST | WIN | BINSEG | OPT | KERNEL | PELT |
|---|---|---|---|---|---|---|---|---|---|---|---|
| 1 | 0.99 | - | - | 0.91 | 0.90 | 1.00 | 0.94 | 0.98 | 0.98 | 0.72 | 0.55 |
| 2 | **1.00** | - | - | 0.81 | 0.97 | 1.00 | 0.75 | 1.00 | 0.84 | 1.00 | 0.89 |
| 3 | 0.98 | - | - | 0.75 | 0.76 | 1.00 | 0.69 | 0.92 | 0.85 | 0.60 | 0.61 |
| 4 | 0.96 | - | - | 0.83 | 0.98 | 1.00 | 0.62 | 0.68 | 0.71 | 0.59 | 0.82 |
| 5 | **0.99** | - | - | 0.68 | 0.62 | 0.46 | 0.50 | 0.50 | 0.64 | 0.57 | 0.73 |
| 6 | **1.00** | - | - | 0.86 | 0.98 | 1.00 | 0.68 | 0.87 | 0.83 | 0.99 | 0.90 |
| 7 | 0.98 | - | - | 0.74 | 0.97 | 1.00 | 0.69 | 0.50 | 0.66 | 0.99 | 0.84 |
| 8 | 0.93 | - | - | 0.71 | 0.59 | 0.67 | 0.76 | 0.50 | 0.59 | 0.62 | 0.70 |
| 9 | 0.97 | 0.98 | 0.80 | 0.89 | 0.74 | 0.77 | 0.68 | 0.68 | 0.61 | 0.59 | 0.57 |
| 10 | **0.88** | 0.52 | 0.73 | 0.62 | 0.64 | 0.57 | 0.79 | 0.74 | 0.79 | 0.84 | 0.79 |
| 11 | **1.00** | - | - | 1.00 | 0.80 | 1.00 | 1.00 | 1.00 | 1.00 | 1.00 | 0.54 |
| 12 | **1.00** | 0.66 | 0.94 | 1.00 | 1.00 | 1.00 | 1.00 | 1.00 | 1.00 | 1.00 | 1.00 |
| 13 | **1.00** | - | - | 1.00 | 0.83 | 1.00 | 1.00 | 1.00 | 1.00 | 1.00 | 1.00 |

**TABLE 14.** RCPD metric for synthetic datasets [21]. The first column contains dataset indices. The header contains algorithm names. In cells, RCPD metric for the corresponding dataset and algorithm is given.

| Dataset name | SDE | ClaSP | BOCPD | KLCPD | TIRE | CHANGE_FOREST | WIN | BINSEG | DYNP | KERNEL | PELT |
|---|---|---|---|---|---|---|---|---|---|---|---|
| 1 | 0.01 | - | - | 0.12 | 0.13 | 0.00 | 0.46 | 0.01 | 0.01 | 0.25 | 0.27 |
| 2 | **0.00** | - | - | 0.16 | 0.01 | 0.00 | 0.33 | 0.00 | 0.16 | 0.00 | 0.06 |
| 3 | 0.01 | - | - | 0.47 | 0.21 | 0.00 | 0.24 | 0.04 | 0.08 | 0.24 | 0.23 |
| 4 | 0.02 | - | - | 0.36 | 0.01 | 0.00 | 0.41 | 0.41 | 0.42 | 0.41 | 0.10 |
| 5 | **0.00** | - | - | 0.20 | 0.19 | 0.22 | 0.31 | 0.26 | 0.07 | 0.11 | 0.21 |
| 6 | **0.00** | - | - | 0.50 | 0.01 | 0.00 | 0.21 | 0.39 | 0.09 | 0.00 | 0.05 |
| 7 | 0.01 | - | - | 0.50 | 0.02 | 0.00 | 0.43 | 0.42 | 0.42 | 0.01 | 0.10 |
| 8 | **0.02** | - | - | 0.23 | 0.17 | 0.25 | 0.09 | 0.26 | 0.21 | 0.08 | 0.12 |
| 9 | 0.02 | 0.01 | 0.48 | 0.50 | 0.20 | 0.13 | 0.18 | 0.18 | 0.41 | 0.33 | 0.33 |
| 10 | 0.01 | 0.00 | 0.03 | 0.00 | 0.05 | 0.05 | 0.02 | 0.04 | 0.02 | 0.02 | 0.02 |
| 11 | **0.00** | - | - | 0.00 | 0.30 | 0.00 | 0.00 | 0.00 | 0.00 | 0.00 | 0.28 |
| 12 | **0.00** | 0.20 | 0.48 | 0.00 | 0.00 | 0.00 | 0.00 | 0.00 | 0.00 | 0.00 | 0.00 |
| 13 | **0.00** | - | - | 0.00 | 0.25 | 0.00 | 0.00 | 0.00 | 0.00 | 0.00 | 0.00 |

We suggest the main reason for such good quality is that our method effectively fits the latent dynamics of a multivariate process with the robust continuous dynamics of SDE. The proposed method generalizes the conventional likelihood ratio CPD methods based on stochastic differential equations [22], [23] with modern DL techniques. To our knowledge, this is the first DL approach that uses Latent Stochastic Differential Equations. Moreover, unlike the preceding SDE-based methods, our algorithm is designed for all main types of change points, including trend, mean, and volatility change.

Along with a good performance, our algorithm has good scalability and computing efficiency. The training and inference stages of the algorithm are linear with the respect to the time series size $N$, whereas lots of other state-of-the-art algorithms with comparable quality are much less scalable, or require much higher computational complexity.

Another main property of our algorithm is flexibility, which helps to use any deep learning based encoder-decoder architectures behind the algorithm. Moreover, more complicated preprocessing techniques can be used, particularly to detect seasonality changes.

This way, we first provide and study a general Neural SDE framework in a CPD setting, which makes it possible to apply a wide range of modern deep learning techniques to CPD problems. This fact, along with all the aforementioned properties, provides a broad perspective for further study and improvements of the proposed algorithm.

## VII. CONCLUSION

The work is aimed to designing an efficient DL generalization of the conventional likelihood ratio CPD approaches based on stochastic differential equations. To this end, we present a first study of Latent SDE in a change point detection setting. As a result of this work, a novel CPD algorithm on the edge of modern deep learning approaches and conventional CPD methods is introduced. This is the first deep learning modification of stochastic differential equations approach to change point detection.

It was theoretically and experimentally shown that the proposed method is capable of detecting all the main types of change points in multivariate time series data: trend, mean, and volatility changes. In most of the scenarios and metrics, the model shows high robustness and a performance which is strongly higher than other state-of-the-art CPD algorithms used in this work.

With all the aforementioned, the proposed algorithm represents a big interest from theoretical and performance perspective for change point detection problem.

## APPENDIX
### A. EXPERIMENT SETUP. TRAIN, INFERENCE, AND IMPLEMENTATION DETAILS

In this auxiliary section, experimental and training specifics details are described.

In the experiments, each algorithm is used with fixed pre-defined hyperparameters. For BINSEG [62], PELT [49], OPT [49], KERNEL [19], and WIN [49], the best values for hyperparameters obtained by the authors of SKAB benchmark were chosen [29]. For BOCPD [17], ClaSP [21], TIRE [24], KLCPD [25], CHANGE_FOREST [36], the best hyperparameters are taken from the corresponding papers.

Neural Stochastic Differential Equation is implemented using torchsde[3] framework with *noise_type = diagonal*, *sde_type = Stratonovich_SDE*. The neural SDE networks configuration is shown in Figure 3. In this work, we use torchsde configuration as a basis for our one [27].

The configuration is trained 100 iterations on each dataset with batch size 512. We monitor the evaluation metrics on each epoch in inference mode and save the best values of it.

---

[3]https://github.com/google-research/torchsde

For training, we use Adam optimizer with learning rate $10^{-2}$ and default parameters.

In all the experiments, we use a machine with single 8-core CPU *Xeon E5-2689* and single *Nvidia 1080 Ti* GPU. The training stage for each dataset on that hardware takes approximately 200 minutes.

### B. SYNTHETIC CORPUS

In our experiments, an additional corpus of synthetic datasets is introduced. We make this corpus to see how each algorithm works on different types of change points. To accomplish that, a set of synthetic tests (datasets) is generated for three main types of change points: trend, mean, and volatility change. In Table 8, a complete set of tests is listed. In Tables 9, 10, 11, 12, 13, 14, a detailed results for NAB.Standart, NAB.LowFP, NAB.LowFN, F1, Covering, and RCPD metrics are shown respectively.

### ACKNOWLEDGMENT

This article follows from Strategic Project "Digital Transformation: Technologies, Effects, and Performance", which is part of Higher School of Economics' development program under the "Priority 2030" academic leadership initiative. The "Priority 2030" initiative is run by Russia's Ministry of Science and Higher Education as part of National Project "Science and Universities".

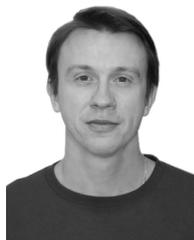

**ARTEM RYZHIKOV** received the M.Sc. degree from the National Research University Higher School of Economics, Russia, where he is currently pursuing the Ph.D. degree in computer science. His current research interests include machine learning and its application to high-energy physics, with a focus on deep learning and Bayesian methods.

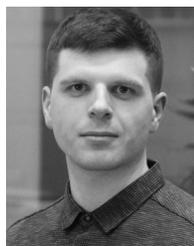

**MIKHAIL HUSHCHYN** received the M.Sc. and Ph.D. degrees in mathematical modeling, numerical methods, and program complexes from the Moscow Institute of Physics and Technology (MIPT), in 2015. He joined HSE University, Moscow, where he is currently a Senior Research Fellow. His current research interests include the application of machine learning and artificial intelligence methods in the natural sciences and industry.

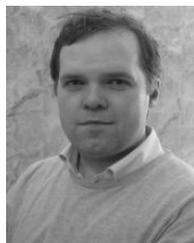

**DENIS DERKACH** received the degree from Saint-Petersburg State University, in 2007, and the Ph.D. degree in particle physics from the University of Paris 11. After postdoctoral training with Istituto Nazionale di Fisica Nucleare and the University of Oxford, he joined the National Research University Higher School of Economics (HSE University), Moscow, where he is currently an Assistant Professor. He is also the Head of the Laboratory of Methods for Big Data Analysis, HSE University. His current research interest includes applying data science methods in the fundamental physics field.

• • •